%% file: main.tex
\documentclass[letterpaper, 10 pt, conference]{ieeeconf}  %
\IEEEoverridecommandlockouts                              %
\include{000_preamble}

\begin{document}

\title{ACRONYM: A Large-Scale Grasp Dataset Based on Simulation}

\author{Clemens Eppner$^{1}$, Arsalan Mousavian$^{1}$, Dieter Fox$^{1,2}$%
\thanks{$^{1}$NVIDIA, USA. $^{2}$University of Washington, Paul G.~Allen School of Computer Science \& Engineering, Seattle, WA, USA. %
{\tt\small ceppner@nvidia.com, amousavian@nvidia.com, dieterf@nvidia.com}}%
}

\maketitle

\begin{abstract}
We introduce ACRONYM, a dataset for robot grasp planning based on physics simulation.
The dataset contains 17.7M parallel-jaw grasps, spanning 8872~objects from 262~different categories, each labeled with the grasp result obtained from a physics simulator.
We show the value of this large and diverse dataset by using it to train a state-of-the-art learning-based grasp planning algorithm.
Grasp performance improves significantly when compared to the original smaller dataset.
Data and tools can be accessed at \url{https://sites.google.com/nvidia.com/graspdataset}.
\end{abstract}

\input{00_introduction.tex}
\input{01_related_work.tex}

\input{02_dataset.tex}
\input{04_graspnet.tex}
\input{05_results.tex}

\input{06_conclusion.tex}

\section*{Acknowledgments}
We thank Miles Macklin, Viktor Makoviychuk, and Nuttapong Chentanez for support with FleX.

\renewcommand*{\bibfont}{\footnotesize}
\printbibliography

\end{document}

%% file: 000_preamble.tex
\usepackage{hyperref}
\usepackage[backend=bibtex,
            hyperref=true,
            url=false,
            isbn=false,
            doi=false,
            backref=false,
            style=ieee,
            natbib=true,%
            mincitenames=1,
            maxcitenames=1,
            citestyle=numeric-comp,
            sorting=nyt,%
            block=none]{biblatex}
\usepackage{siunitx}
\usepackage{graphicx}

\usepackage{manfnt}
\usepackage{amssymb}
\usepackage{bbding}
\usepackage{fontawesome}
\usepackage{dsfont}

\usepackage{makecell}
\usepackage{color, colortbl}
\newcolumntype{R}[2]{%
    >{\adjustbox{angle=#1,lap=\width-(#2)}\bgroup}%
    l%
    <{\egroup}%
}

\newcommand{\rot}[1]{\rotatebox{90}{#1}}

\usepackage{subcaption}

\newcommand\mypara[1]{\vspace{0mm}\textbf{#1}}

\definecolor{LightGray}{gray}{0.9}

\newcommand\Tstrut{\rule{0pt}{2.6ex}}         %
\newcommand\Bstrut{\rule[-0.9ex]{0pt}{0pt}}   %

%% file: 00_introduction.tex
\section{Introduction}
Devising algorithms that exhibit competent grasping and manipulation skills is a core robotics research question. Over the past years, a surge of methods has been presented that tackle the grasping problem in a data-driven manner~\cite{lenz2015deep,Levine2016,Mahler2017,ten2017grasp,tobin2018domain}.
The success of these approaches critically depends on the data that is being used.

There are three fundamental sources for grasping data: real-world trial-and-error~\cite{pinto2016supersizing,Levine2016}, human annotation~\cite{Kappler2015,yan2017learning}, and synthetically generated grasp labels. Among the last category one can distinguish between analytical grasp models~\cite{Mahler2017} and full-fledged physics simulations~\cite{Kappler2015,depierre2018jacquard}.
In this paper, we focus on arguably the most scalable alternative: generating synthetic grasps with a physics simulator.

Given the range of existing grasp datasets~(see Table~\ref{tab:datasets}), one might ask: Why yet another one? 
Our dataset is unique in the sense that it combines a number of desirable properties:
\\\textbf{3D:} The current landscape of learning-based grasping approaches is dominated by algorithms that work with planar grasp representations~\cite{jiang2011efficient,lenz2015deep,redmon2015real,wang2016robot,asif2017rgb,guo2017hybrid,kumra2017robotic,asif2018graspnet,ghazaei2018dealing,chu2018real,zhou2018fully}. In contrast, our dataset allows for spatial grasping, which is especially useful in constrained spaces and when semantics become important.
\\\textbf{High Grasp Volume and Density:}
We provide a rather high density of grasps per object compared to existing datasets~(see~Sec.~\ref{sec:related_work}). This is important when trying to learn grasp manifolds.
\\\textbf{Physical Realism:} Synthetically generated grasp data often uses analytical measures to label success~\cite{goldfeder2008columbia,Mahler2017}. It has been shown that such measures not always transfer to the real world. Instead, our dataset uses the physics simulator FleX~\cite{macklin2014unified} to label grasps. Although being computationally more expensive, simulation has been shown to resemble real-world grasp performance much more closely~\cite{danielczuk2019reach,mousavian20196}. 
\\\textbf{Scene Variation:} Aside from single-object scenes we also provide cluttered scenarios containing multiple objects on a support surface.
This information can be used to learn e.g. collision-free grasps in cluttered scenarios~\cite{murali2019cluttergrasping}.

\begin{figure}
    \centering
    \includegraphics[width=\linewidth]{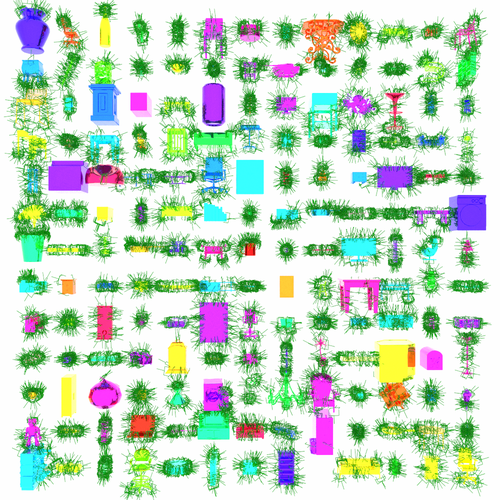}
    \caption{ACRONYM contains 2000~parallel-jaw grasps for 8872~objects from 262~categories, totalling 17.7M grasps.}
    \label{fig:intro_dataset}
\end{figure}

In conclusion, this paper contains two main contributions:
\begin{enumerate}
    \item We provide a large-scale grasp dataset, containing a total of \SI{17.744}{} million parallel-jaw grasps for \SI{8872}{} different objects from ShapeNetSem~\cite{savva2015semantically}. The grasps are labeled using a physics simulation.
    \item We show that re-training a state-of-the-art learning-based grasp planner~\cite{mousavian20196} on our dataset leads to higher performance compared to the original dataset. This is due to the greater variability offered by ACRONYM.
\end{enumerate}

The rest of the paper is organized as follows: After reviewing and contrasting related work, we describe the data generation process in detail.
The dataset and accompanying tools are available at~\url{https://sites.google.com/nvidia.com/graspdataset}.

%% file: 01_related_work.tex
\section{Related Work}
\label{sec:related_work}

\begin{table*}[!thbp]
    \centering
    \begin{tabular}{|c|c|c|c|c|c|c|c|c|}\hline
        Dataset & \rot{\small Planar/3D~} & \rot{\small \parbox{1cm}{Observa-tions}} & \rot{\small Labels} & \rot{\small Grasps} & \rot{\small \parbox{1cm}{Objects\\(Cat.)}} & \rot{\small \parbox{1cm}{Grasps per Obj./Sce.}} & \rot{\small Scenes} & \rot{\small Used by} \Bstrut\\\hline\hline
        \small{Cornell~\cite{jiang2011efficient}} & $\square$ & Real & \HandPencilLeft & 8k & 240 & 33 & \small Single & \makecell{\tiny \cite{jiang2011efficient,lenz2015deep,redmon2015real,wang2016robot}\\ \tiny \cite{asif2017rgb,guo2017hybrid,kumra2017robotic,asif2018graspnet}\\ \tiny \cite{ghazaei2018dealing,chu2018real,zhou2018fully}} \Tstrut\\\rowcolor{LightGray}
        \small{Jacquard~\cite{depierre2018jacquard}} & $\square$ & Sim & \faPlayCircle & 1.1M & 11k & 100 & \small Single & \small \cite{depierre2018jacquard,zhou2018fully,zhang2018roi} \\
        \small{VMRD + Grasps~\cite{zhang2018roi}} &  $\square$ & Real & \HandPencilLeft & 100k & $\approx$15k (31) & $\approx$6.5 & \small Multi & \cite{zhang2018roi}\\\rowcolor{LightGray}
        \small{Levine et al.~\cite{Levine2016}} & $\square$ & Real & \faEye & 650k & N/A & N/A & \small Bin & \cite{Levine2016} \\
        \small{Columbia~\cite{goldfeder2008columbia}} & \mancube & Sim & $f$ & 238k & 7256 (161) & $\approx$32 & \small Single & \cite{goldfeder2009data}\\\rowcolor{LightGray}
        \small{Kappler et al.~\cite{Kappler2015}} & \mancube & Sim & \small{\HandPencilLeft}, \small{\faPlayCircle} & 300k & 700 (80) & $\approx430$ & \small Single & \cite{Kappler2015} \\
        \small{Dex-Net~\cite{Mahler2017}} & \mancube & Sim & $f$ & 6.7M & 1500 (50) & 100 & & \cite{Mahler2017} \\\rowcolor{LightGray}
        \small{Veres et al.~\cite{veres2017integrated}} & \mancube & Sim & \faPlayCircle & 50k & N/A (64) & N/A & \small Single & \\
        \small{6-DOF GraspNet~\cite{mousavian20196}} & \mancube & Sim & \faPlayCircle & 7.07M & 206 (6) & 34k & \small Single &   \cite{mousavian20196} \\\rowcolor{LightGray}
        \small{Eppner et al.~\cite{eppner2019graspsampling}} & \mancube & - & \faPlayCircle & 1B & 21 & 47.8M & \small Single &  \\
        \small{GraspNet~\cite{fang2020graspnet}}  & \mancube & S+R & $f$ & 1.1B & 88 & 12.5M & \small Multi & \\\rowcolor{LightGray}
        \small{{EGAD!}~\cite{morrison2020egad}} & \mancube & Sim & $f$ & 233k & 2331 & 100 & Single & \\
        \textbf{ACRONYM} & \mancube & Sim & \faPlayCircle & 17.7M & 8872 (262) & 2000 & \small Multi & \Bstrut\\\rowcolor{LightGray}\hline
    \end{tabular}
    \caption{Comparison of publicly available grasp datasets. Grasp labels are generated either manually~(\HandPencilLeft), by physics simulation~(\faPlayCircle), analytical models~($f$), or through observation of real-world trials~(\faEye).}
    \label{tab:datasets}
\end{table*}

Existing grasp datasets differ in the type of observations, how grasp labels are produced, and the amount of variability they contain~(see Table~\ref{tab:datasets}).
We focus our discussion on publicly available datasets.
Note, that we do not include datasets that are based on human grasping and manipulation~\cite{huang2016recent} since it is non-trivial to adapt them to robotic use cases.

\textbf{Planar vs. Spatial Grasping Datasets.}
A significant amount of data-driven grasp approaches search for planar grasps, i.e., grasps are elements of $SE(2)$. This usually means that the gripper is aligned with the image plane, as popularized by the rectangular grasp representation~\cite{jiang2011efficient}.
This includes the Cornell Grasping Dataset~\cite{jiang2011efficient}, the data from~\citet{Levine2016}, the extension of the VMRD~\cite{zhang2018roi}, and the Jacquard dataset~\cite{depierre2018jacquard}.
Our dataset provides full 6DOF grasp poses similar to~\cite{goldfeder2008columbia,Kappler2015,Mahler2017,veres2017integrated}.

\textbf{Synthetic vs. Real Observations.}
Datasets that provide real RGB-D measurements are either limited in scale~\cite{jiang2011efficient} or limited to planar grasping~\cite{jiang2011efficient,Levine2016,zhang2018roi}. Simulating observations provides a more scalable alternative. Synthetic datasets contain either depth images \cite{Kappler2015,Mahler2017} or RGB-D~\cite{veres2017integrated,depierre2018jacquard}.
The Columbia grasp database~\cite{goldfeder2008columbia} itself does not contain any observational data but can be combined with any off-the-shelf renderer.

\textbf{Grasp Labels.}
The main differentiator between the datasets is the way grasps are labeled.
While some rely on real-world robot executions~\cite{Levine2016}, many use analytical models~\cite{goldfeder2008columbia,Mahler2017,fang2020graspnet} or physics simulators~\cite{Kappler2015,veres2017integrated,depierre2018jacquard}.
A few even contain human annotated grasp data~\cite{jiang2011efficient,Kappler2015}.

\textbf{Variety of Scenes.}
Most grasp datasets provide grasp labels for single objects~\cite{jiang2011efficient,depierre2018jacquard,goldfeder2008columbia,mousavian20196}, rather than cluttered scenes with multiple objects~\cite{zhang2018roi,Levine2016,fang2020graspnet}.
Our dataset contains grasp data for single objects but also provides a mechanism to generate scenes with multiple objects.
In this case, single-object grasp labels are reused with colliding grasps being ignored.

\textbf{Quantity.}
The amount of grasp data depends on the number objects/scenes and the grasp density for each of them.
Some datasets focus on a larger variety of objects or scenes~\cite{goldfeder2008columbia,zhang2018roi,Levine2016} while others provide lots of grasps for few objects~\cite{fang2020graspnet,eppner2019graspsampling}.
It is still unclear which aspect is more important for generalized grasping.
But it has been shown that geometric variability can be captured by randomly generated shapes~\cite{tobin2018domain}.
We try to balance the need for variability by providing data for many objects~(8872) with rather high grasp density per object~(2000).
In total, this amounts to \SI{17.744}{} million grasps. %

%% file: 02_dataset.tex
\section{Dataset Generation}
\label{sec:data_set}
In the following, we describe our process of generating grasps in simulation and scenes with structured clutter.

\mypara{Objects.}
We chose ShapeNetSem~\cite{savva2015semantically} as a source for object meshes.
Although it has been shown that random geometries can be used to learn grasp estimators~\cite{tobin2018domain}, using semantically meaningful meshes will help to tackle task-specific grasping problems. 
\begin{figure}
    \centering
    \begin{subfigure}[b]{1.\linewidth}
        \centering
        \includegraphics[width=.9\textwidth]{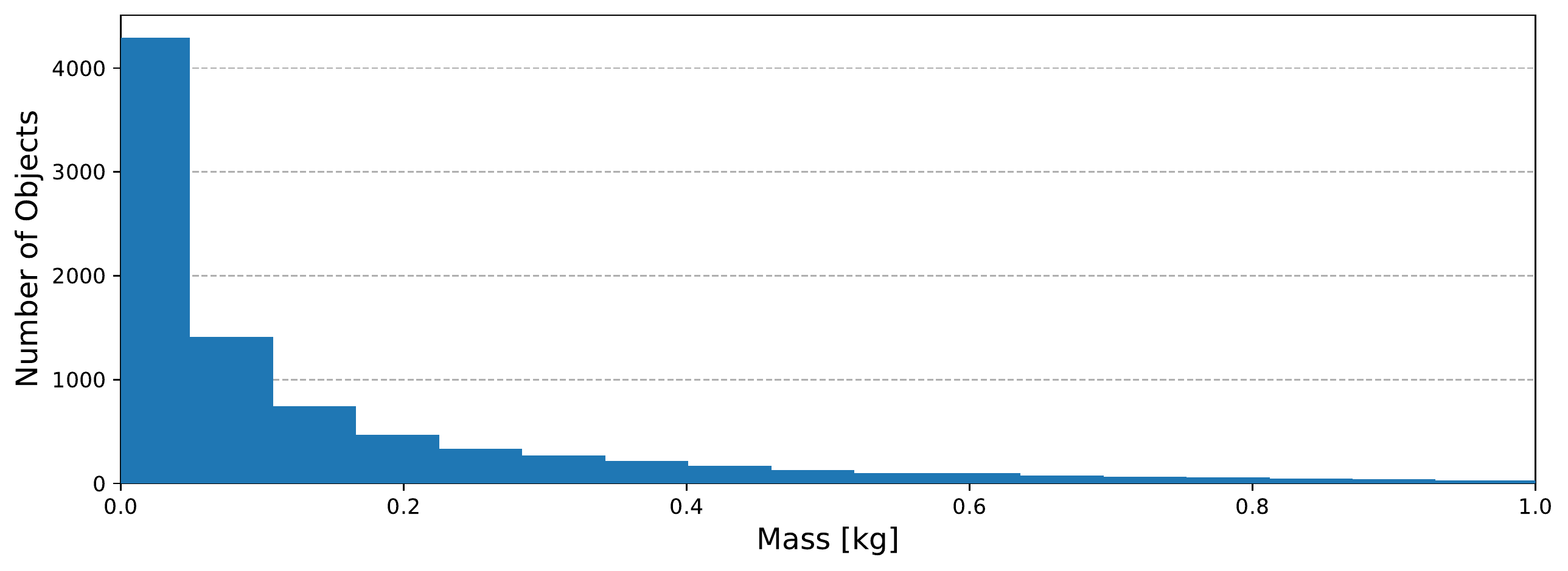}
    \end{subfigure}\\%
    \begin{subfigure}[b]{1.\linewidth}
        \centering
        \includegraphics[width=.9\textwidth]{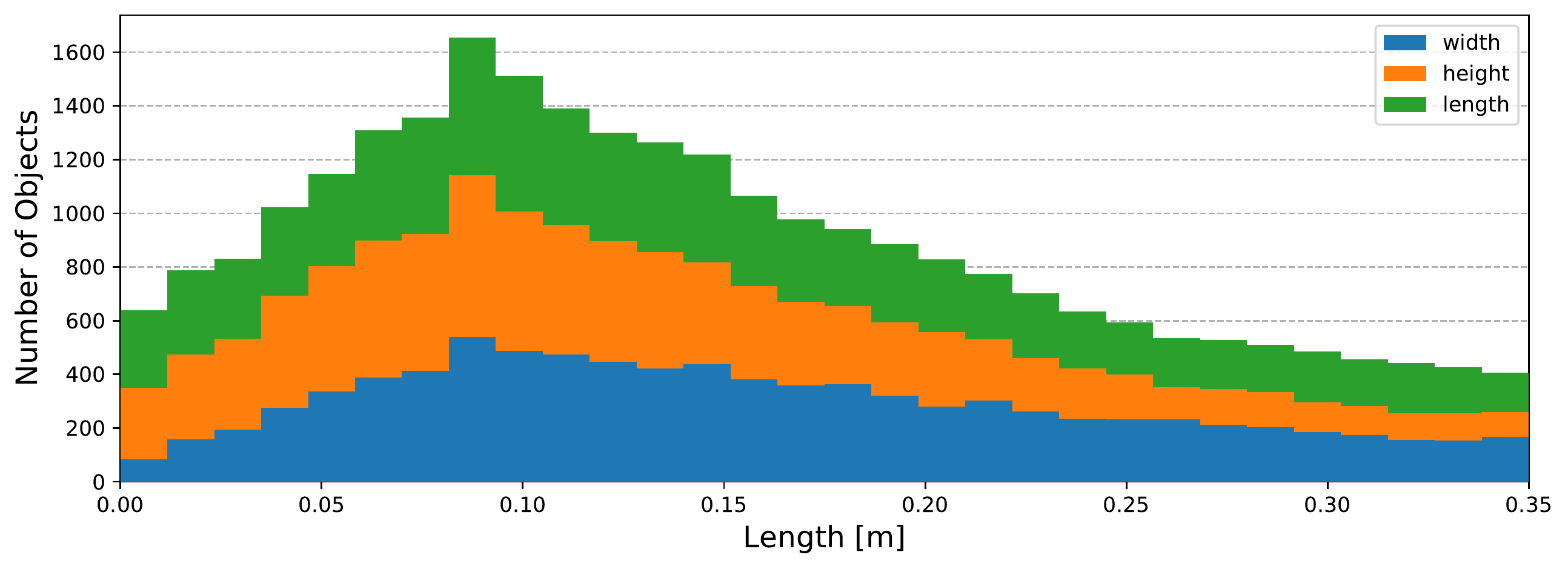}
    \end{subfigure}%
    \caption{Distribution of object masses and dimensions in the dataset.}
    \label{fig:object stats}
\end{figure}

Since ShapeNet is a well-established object dataset it also simplifies exploiting closely related research results in pose estimation, shape completion, semantic segmentation etc.
We exclude meshes that contain more than one connected component.
To ensure that the meshes behave correctly during physics simulation, we make them watertight~\cite{huang2018robust}.
We assume uniform density, i.e., center of mass and geometric center of the objects coincide.
All objects exhibit the same friction~(\SI{1.0}{}) and density~(\SI{150}{\kg\per\m^{3}}).
We sample object scale such that the longest side of an axis-aligned bounding box for each object is uniformly distributed between \SI{6}{\cm} and \SI{35}{\cm}.
The resulting mass and dimension distributions are shown in Fig.~\ref{fig:object stats}.

\mypara{Grasp Sampling.}
Grasps in our dataset are parameterized by $g = (g_{e}, g_{i})$, where $g_{e} \in SE(3)$ are the external DOF and $g_{i} \in \mathds{R}^{n}$ the internal DOF of the hand.
Our dataset focuses on parallel-jaw grippers~($n = 1$), specifically it uses the model of the Franka Panda gripper~\cite{emika2018panda}.
Strictly speaking, $g$ defines a pre-grasp which might lead to a stable grasp when closing the fingers.
In the context of the dataset, we use the terms grasp and pre-grasp interchangebly.
All pre-grasps in the dataset have the Panda's maximum grip aperture of~\SI{8}{\cm}.

Given an object mesh we use an antipodal sampling scheme. It consists of sampling an arbitrary point on the mesh surface and a line sampled within the cone aligned with the surface normal. The intersection of the line with the mesh defines the second contact point. The gripper pose $g_{e}$ is then derived by taking the center point between both contacts and a uniformly sampled rotation around the line.
It has been empirically shown that an antipodal sampling creates the most diverse grasps in the range of up to~\SI{100}{k}~grasps per object, compared to other sampling schemes~\cite{eppner2019graspsampling}.

We apply the commonly used heuristic that grasps are unsuccessful if object and hand collide or if the intersection between object and the volume between the gripper fingers is empty.
We generate \SI{2000}{}~grasp proposals per object that pass this test using rejection sampling.

\mypara{Grasp Labelling in Simulation.}
We use the physics simulator FleX~\cite{macklin2014unified} to evaluate and label each grasp.
Objects, gripper palm and fingers are simulated as rigid bodies; gravity is not present.
\begin{figure}
    \centering
    \includegraphics[width=\linewidth]{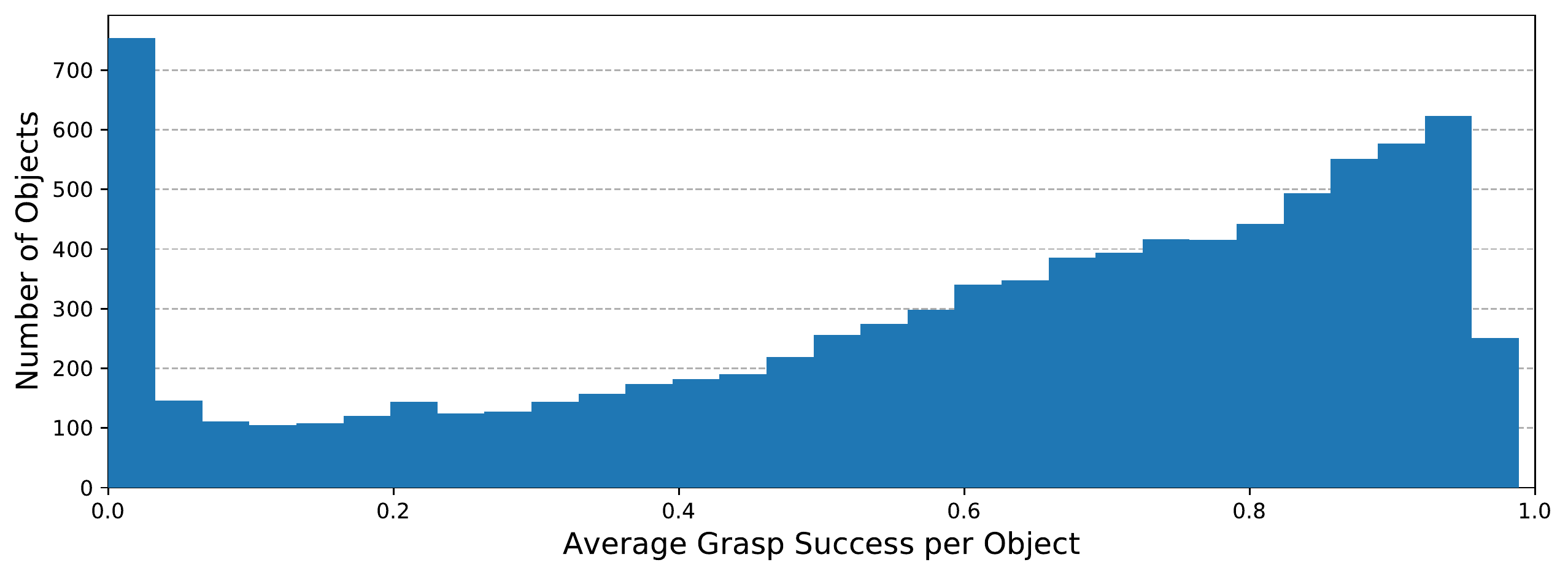}\\%
    \caption{Distribution of grasp successes in the dataset.}
    \label{fig:grasp successes per object}%
\end{figure}
The simulation is initialized with the gripper in the pre-grasp state.
The gripper itself is simulated as an unconstrained position-controlled object.
Subsequently, fingers are closed using a velocity-based controller until a force threshold is reached or the hand is fully closed.
Finally, a shaking motion is executed: The hand first moves up and down along its approach direction, then rotates around a line parallel to the prismatic joint axes of the fingers.
Afterwards, we record grasp success by testing whether the object is still in contact with both fingers.

Shaking applies disturbances in all directions without assuming a particular gravity direction, which is suitable for 6-DOF grasping. The same gravity-less shaking procedure was used in~\cite{Kappler2015,zhou20176dof}.  Additionally, we conducted an experiment to compare grasp results with and without gravity.  Our experiments showed that shaking the object has 91\% precision and 93\% recall on the successful grasps with gravity. This shows that they are almost equivalent to each other, while shaking induces less bias.

\begin{figure}
    \includegraphics[width=.5\linewidth]{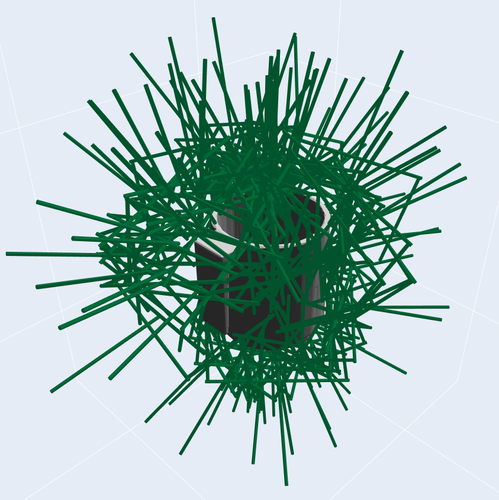}%
    \includegraphics[width=.5\linewidth]{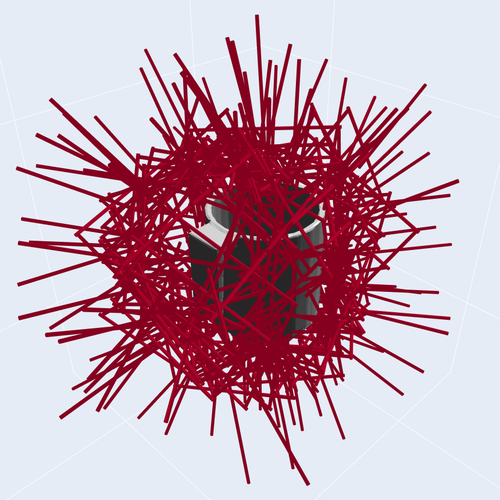}%
    \caption{Green markers show successfully simulated grasps, failed ones are red. Each object contains 2000 grasps. For visualization purposes both images show only \SI{15}{\%} of all labels.}
    \label{fig:grasp examples}
\end{figure}

Fig.~\ref{fig:grasp examples} shows an example result of a mug.
In total, we simulate \SI{17.744} million grasps, out of which \SI{59.21}{\%}~(approximately \SI{10.5} million grasps) succeed.
The distribution of object-wise success rates is shown in Fig.~\ref{fig:grasp successes per object}.

\begin{figure*}[!tbp]
    \centering
    \includegraphics[width=.125\linewidth]{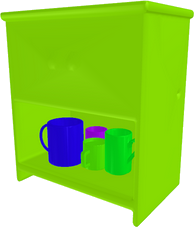}\hfill%
    \includegraphics[width=.125\linewidth]{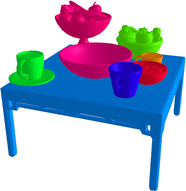}\hfill%
    \includegraphics[width=.125\linewidth]{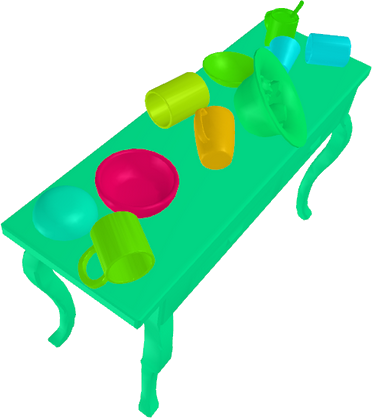}\hfill%
    \includegraphics[width=.125\linewidth]{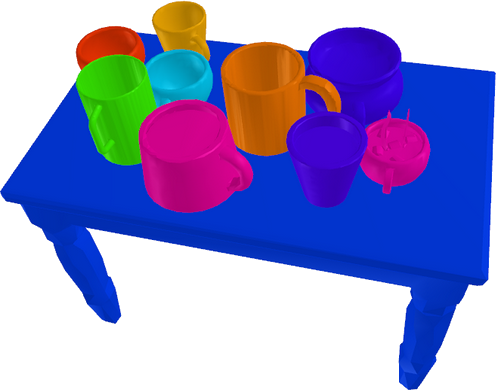}\hfill%
    \includegraphics[width=.125\linewidth]{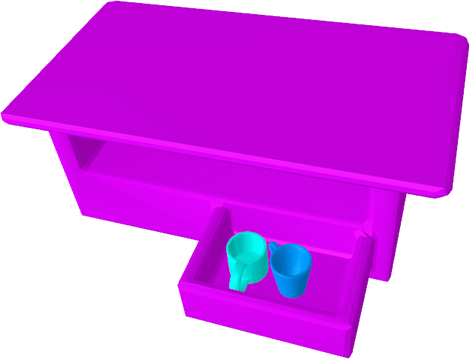}\hfill%
    \includegraphics[width=.125\linewidth]{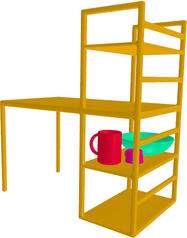}%
    \caption{Grasping scenes with structured clutter. The dataset entails a mechanism to randomly generate scenes with objects clustered on support surfaces of arbitrary meshes. Colors are for visualization purposes only.}
    \label{fig:sim_and_clutter}
\end{figure*}

\mypara{Scenes with Structured Clutter}
In addition to single-object scenes our dataset contains a mechanism to procedurally generate scenes with structured clutter, i.e., multiple objects placed on top of a support surface.
We generate these scenes by sampling a support object (e.g. from ShapeNetSem's furniture class) and a mesh facet whose normal is aligned with the gravity vector and whose area is larger than a minimum threshold.
Given this support surface we sequentially sample locations from a 2D Gaussian (centered around the surface polygon's centroid) to place arbitrary objects on top of it, while ensuring that the resulting configuration is collision-free and the projected center of masses inside the support polygon.
Object orientations can be either sampled from a set of pre-calculated stable poses (using trimesh\footnote{\url{https://github.com/mikedh/trimesh}}) or using the semantically more meaningful up-vector provided by ShapeNetSem.
Examples of resulting scenes are shown in Fig.~\ref{fig:sim_and_clutter}.

To label these scenes with grasps we reuse the grasp labels obtained by simulating single objects.
In addition, we label those grasps as failures that would collide with any scene geometry.
Note, that this is a simplification since environmental contacts could lead to false negatives as well as false positives.
Still, given the type of grasp strategies we are focusing on~(static pre-grasp + finger closing), this seems like a reasonable assumption.

\mypara{Observations}
We do not provide images explicitly.
Instead we provide code based on pyrender\footnote{\url{https://github.com/mmatl/pyrender}} to render depth images, segmenation masks, and point clouds.
Note that ShapeNet meshes do not contain textures but other methods can be used to render realistic materials~\cite{park2019photoshape}.

%% file: 04_graspnet.tex
\section{Baseline Methods: GraspNet and GPD}
\begin{figure*}
    \centering
    \begin{subfigure}[b]{.5\linewidth}%
        \includegraphics[width=\linewidth]{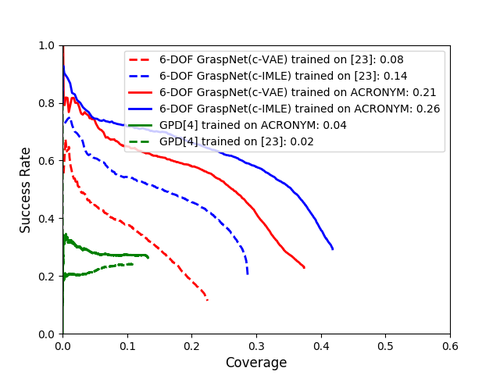}%
        \caption{Evaluation on ACRONYM}
    \end{subfigure}%
    \begin{subfigure}[b]{.5\linewidth}%
        \includegraphics[width=\linewidth]{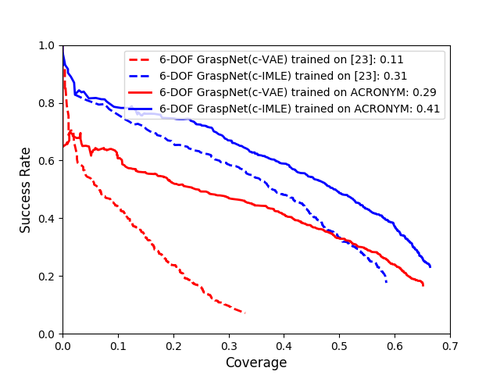}%
        \caption{Evaluation on dataset~\cite{mousavian20196}}
    \end{subfigure}%
\caption{Effect of training data size on generalization and accuracy: A higher area under curve (see legends) indicates better grasp performance. Models trained with the same architecture are in the same colors. The performance consistently improves by training on the ACRONYM dataset. (a) captures generalization of the model while (b) evaluates the models on unseen instances of the categories that all models have seen during training.}
\label{fig:dataset_results}
\end{figure*}

We use our dataset to train 6-DOF GraspNet~\cite{mousavian20196}\footnote{\url{https://github.com/NVlabs/6-DOF-graspnet}} and GPD~(Grasp Pose Detection)~\cite{ten2017grasp}\footnote{\url{https://github.com/atenpas/gpd}}. These two learning-based methods generate a set of grasps given an object point cloud.
6-DOF GraspNet consists of a grasp sampling network, a grasp evaluator network, and a refinement procedure which uses the grasp evaluator to iteratively improve the predicted grasps. The conditional Variational Auto-encoder (c-VAE) was used in~\cite{mousavian20196}.
In addition, we also consider another formulation using conditional Implicit Maximum Likelihood Estimation~(c-IMLE)~\cite{CIMLE}.
c-IMLE addresses the problem of mode collapse~\cite{Bau_2019_ICCV}, which GANs are notorious for, by maximizing the likelihood of each ground truth data point instead of having a discriminator model that tries to distinguish fake from real examples.
More concretely, given the point cloud of an object at a fixed random view point and uniformly sampled latents, the objective is to minimize the distance of each ground truth grasp for that object to any of the generated grasps. The loss objective for c-IMLE is as follows:
$\mathcal{L}_{\text{c-IMLE}} = \sum_{g \in G^*} \min d(g, \hat{g})$
where $g$ represents sampled grasp from the set of ground truth grasps $G^*$ for the object, $\hat{g}$ is the predicted grasp given the object point cloud and the sampled latent, and $d(\cdot,\cdot)$ is the distance function between the grasps that was used in~\citet{mousavian20196}.

GPD~\cite{ten2017grasp} represents grasps by using three local image projections of the volume between both fingers. These images are classified with a four-layer CNN.
GPD is trained by pre-generating 5 views and 500 grasps per object. We used the 15-channels settings of GPD as it is suggested to give the best performance. We trained the model following the official parameters. Evaluation was done by sampling 100~grasps for each point cloud.

%% file: 05_results.tex
\section{Experiments and Results}
\label{sec:baselines}

We evaluate the performance of the different variations of the model on the evaluation set of the ACRONYM dataset and the dataset of~\citep{mousavian20196}.
Here, we focus on the success-coverage curve as the main metric.
The evaluation data is generated by uniformly sampling categories and then choosing a random object from held-out instances of that category.
Once the object instance is selected, it is rendered from a random view point.
The evaluation data for each dataset is precomputed and kept fixed to make sure all methods are evaluated on the exact same objects and viewpoints, and there are no other contributing effects to the difference between different variations.

\begin{figure*}
    \centering
    \begin{subfigure}[b]{.1\linewidth}
         \includegraphics[width=\linewidth]{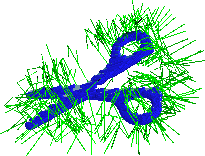}
    \end{subfigure}%
    \begin{subfigure}[b]{.1\linewidth}
         \includegraphics[width=\linewidth]{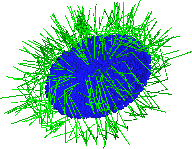}%
    \end{subfigure}%
    \begin{subfigure}[b]{.1\linewidth}
        \includegraphics[width=\linewidth]{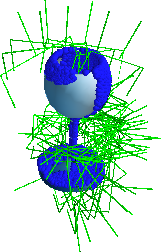}%
    \end{subfigure}%
    \begin{subfigure}[b]{.1\linewidth}
        \includegraphics[width=\linewidth]{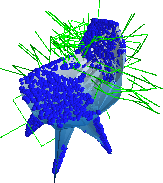}%
    \end{subfigure}%
    \begin{subfigure}[b]{.1\linewidth}
        \includegraphics[width=\linewidth]{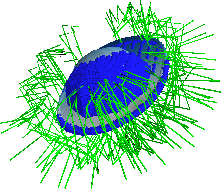}%
    \end{subfigure}%
    \begin{subfigure}[b]{.1\linewidth}
         \includegraphics[width=\linewidth]{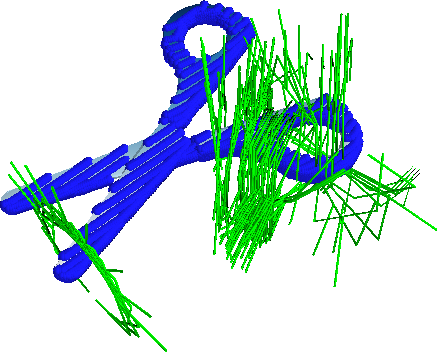}
    \end{subfigure}%
    \begin{subfigure}[b]{.1\linewidth}
             \includegraphics[width=\linewidth]{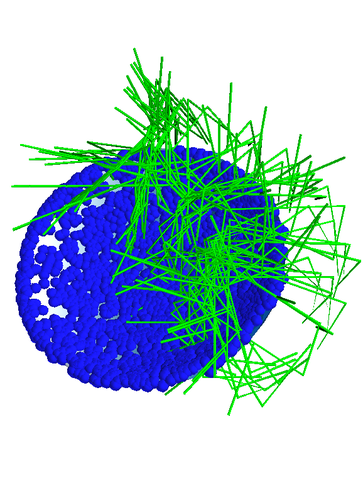}%
    \end{subfigure}%
    \begin{subfigure}[b]{.1\linewidth}
        \includegraphics[width=\linewidth]{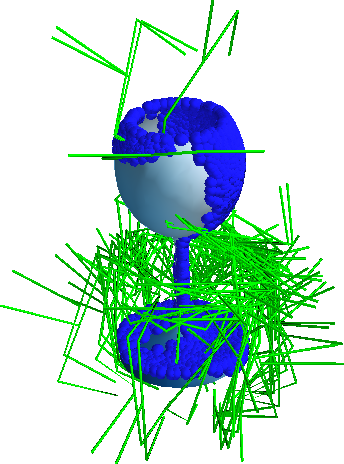}%
    \end{subfigure}%
    \begin{subfigure}[b]{.1\linewidth}
        \includegraphics[width=\linewidth]{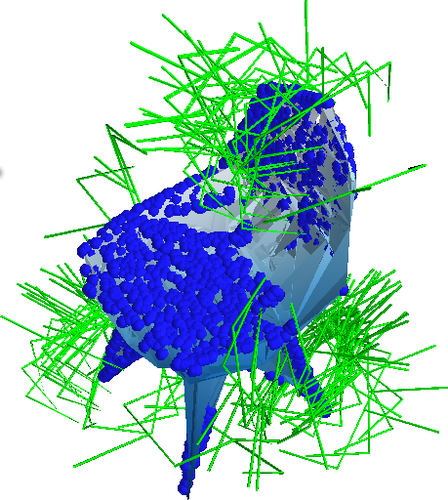}%
    \end{subfigure}%
    \begin{subfigure}[b]{.1\linewidth}
        \includegraphics[width=\linewidth]{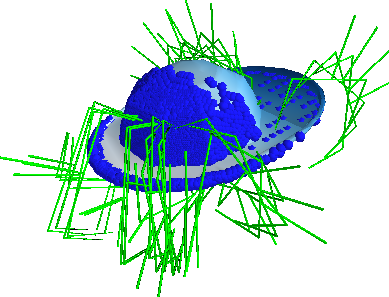}%
    \end{subfigure}\\%
    \begin{subfigure}[b]{.1\linewidth}
        \includegraphics[width=\linewidth]{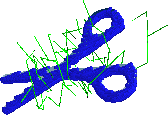}%
    \end{subfigure}%
    \begin{subfigure}[b]{.1\linewidth}
        \includegraphics[width=\linewidth]{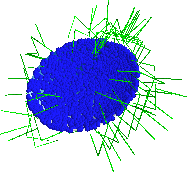}%
    \end{subfigure}%
    \begin{subfigure}[b]{.1\linewidth}
        \includegraphics[width=\linewidth]{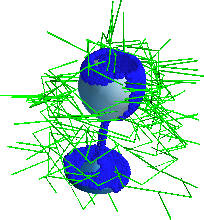}%
    \end{subfigure}%
    \begin{subfigure}[b]{.1\linewidth}
        \includegraphics[width=\linewidth]{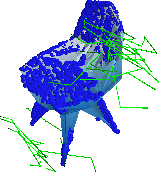}%
    \end{subfigure}%
    \begin{subfigure}[b]{.1\linewidth}
        \includegraphics[width=\linewidth]{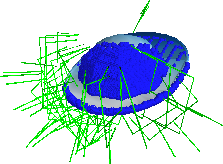}%
    \end{subfigure}%
    \begin{subfigure}[b]{.1\linewidth}
        \includegraphics[width=\linewidth]{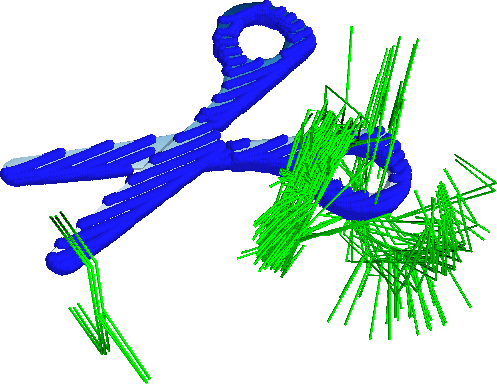}%
    \end{subfigure}%
    \begin{subfigure}[b]{.1\linewidth}
        \includegraphics[width=\linewidth]{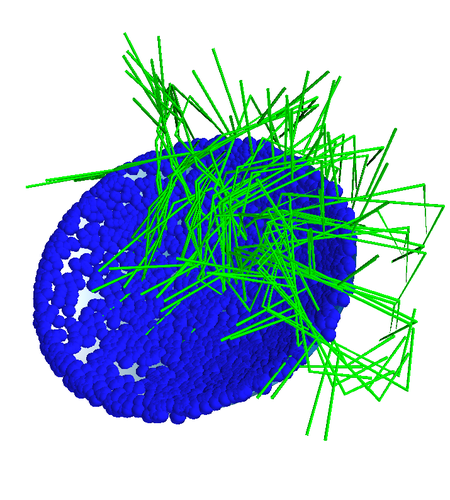}%
    \end{subfigure}%
    \begin{subfigure}[b]{.1\linewidth}
        \includegraphics[width=\linewidth]{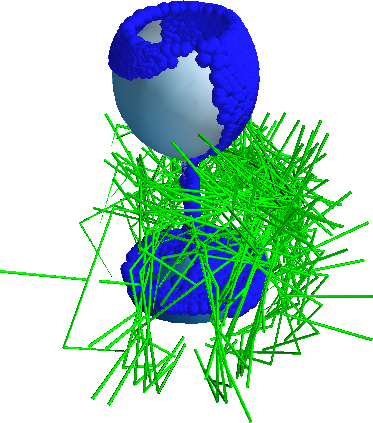}%
    \end{subfigure}%
    \begin{subfigure}[b]{.1\linewidth}
        \includegraphics[width=\linewidth]{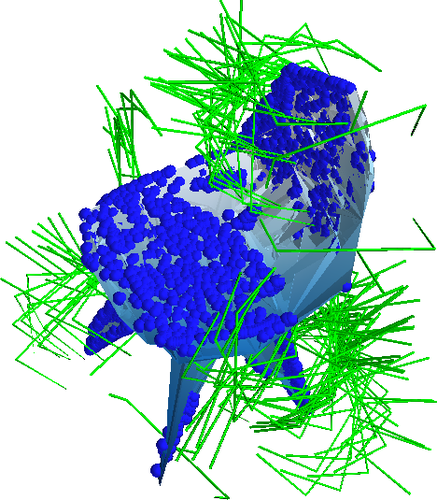}%
    \end{subfigure}%
    \begin{subfigure}[b]{.1\linewidth}
        \includegraphics[width=\linewidth]{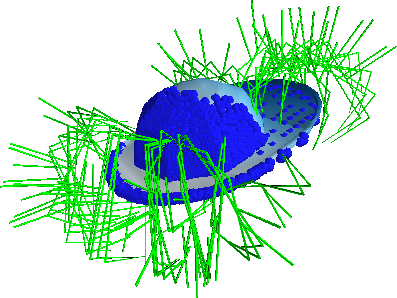}%
    \end{subfigure}%
    \caption{Qualitative comparison of two state-of-the-art grasp planning methods trained on ACRONYM~(\textit{top row}) vs. the smaller, less diverse dataset of~\cite{mousavian20196} ~(\textit{bottom row}). The left half shows grasps generated by 6-DOF GraspNet (c-IMLE)~\cite{mousavian20196} while the right one shows GPD~\cite{ten2017grasp}. Blue dots indicate observed points. The models trained on ACRONYM generate more diverse and robust grasps over all six example objects, as shown by the green markers.
    }
    \label{fig:qualitative}
\end{figure*}

{\bf Does more data lead to more generalization on unseen categories?}
The performance of data-driven methods usually increases with the size of the training set.
However, at some point the performance saturates.
We want to investigate how ACRONYM can increase the performance of our baselines with the exact same model capacity.
It is possible to achieve better performance with larger models on the bigger dataset, but to clarify the benefit of ACRONYM we kept all model parameters identical and only changed the training data.
We compared six combinations: c-VAE and c-IMLE variations of 6-DOF GraspNet and GPD that are trained either on the dataset of~\cite{mousavian20196} or ACRONYM. The dataset of~\cite{mousavian20196} is a subset of the ACRONYM dataset and as a result models trained on~\cite{mousavian20196} are evaluated on their generalization capability on unseen categories.
Fig~\ref{fig:dataset_results}-a shows that the c-IMLE variation consistently outperforms the c-VAE variation and GPD. And more importantly, the performance of all three models increases by 2-3 fold when trained on ACRONYM. These results show the necessity of large-scale datasets to increase the generalization of grasping methods. Fig.~\ref{fig:qualitative} shows the difference between the models trained on ACRONYM dataset and the ones trained on \cite{mousavian20196}. For most of the objects in Fig.~\ref{fig:qualitative} the difference is significant. Note, that for the wine mug the model trained on~\cite{mousavian20196} generates many unsuccessful but highly confident grasps.
Most of the failures that GPD produces result from the occlusion boundary of the point cloud. These failures are due to processing only points between the fingers and not considering the global context of the object.

{\bf Does more data lead to better grasps on seen categories but unseen instances?} In the previous section, one could argue that the models that are trained with the training data of~\cite{mousavian20196} are at a disadvantage because the model is trained on fewer categories and naturally would be inferior. To address this concern, we compared the models on the held-out objects of~\cite{mousavian20196}. The dataset contains meshes from the mug, bowl, bottle, cylinder, and box categories. The models trained on~\cite{mousavian20196} are trained on all the evaluation categories. However, the model that is trained on ACRONYM is not trained on any boxes or cylinders. Note that all of the models have the exact same number of parameters and the only difference is that one is trained on more categories and instances while the other variation is trained on a limited number of categories. Fig.~\ref{fig:dataset_results}-b shows that the model trained on ACRONYM outperforms the other even on those categories. These results show the necessity of having a large-scale dataset such as ACRONYM. It can help improve the grasping capability of data-driven methods. In addition, it is worth noting that all the methods consistently perform significantly worse on ACRONYM which shows that the data has more depth to be a sustainable long-term benchmark.

%% file: 06_conclusion.tex
\section{Limitations}
\label{sec:limitations}
One might question the quality of the dataset, since it was generated purely in simulation.
We argue that grasp simulation has been successfully transferred from simulation to the real world in the past, in planning~\cite{dogar2012physics} as well as learning contexts~\cite{tobin2018domain,depierre2018jacquard,mousavian20196}.
In particular, grasps successfully simulated with the physics engine FleX~(the same we use) have been reproduced in the real world~\cite{danielczuk2019reach,eppner2019graspsampling,murali2019cluttergrasping}.
\citet{danielczuk2019reach} execute \SI{2625}{} grasps on a real robot and show that FleX has an average precision of~\SI{86}{\%}, the highest among all compared models.
Finally, \citet{Kappler2015} used crowdsourced data to show that simulation-based metrics are more predictive of grasp success than their analytical counterparts~(which are used in other large-scale datasets~\cite{Mahler2017,fang2020graspnet}).

\section{Conclusion}
\label{sec:conclusion}
In this paper, we introduced a new large-scale grasp dataset based on physics simulation called ACRONYM.
We showed its usefulness by using it to re-train two existing learning-based grasp methods. 
As a result, the methods not only perform better on unseen objects of ACRONYM but also generalize better to unseen objects of the original, much smaller dataset.
We hope that the introduction of the ACRONYM dataset will help robotics researchers to innovate new algorithms and contrast existing ones.